# Real-time Noise Detection and Classification in Single-Channel EEG: A Lightweight Machine Learning Approach for EMG, White Noise, and EOG Artifacts

Hossein Enshaei [a], Pariya Jebreili [a], Sayed Mahmoud Sakhaei [a]*

[a] Department of Electrical and Computer Engineering, Babol Noshirvani University of Technology, Babol, Iran

## Abstract

Electroencephalogram (EEG) artifact detection in real-world settings faces significant challenges such as computational inefficiency in multi-channel methods, poor robustness to simultaneous noise, and trade-offs between accuracy and complexity in deep learning models. We propose a hybrid spectral-temporal framework for real-time detection and classification of ocular (EOG), muscular (EMG), and white noise artifacts in single-channel EEG. This method, in contrast to other approaches, combines time-domain low-pass filtering (targeting low-frequency EOG) and frequency-domain power spectral density (PSD) analysis (capturing broad-spectrum EMG), followed by PCA-optimized feature fusion to minimize redundancy while preserving discriminative information. This feature engineering strategy allows a lightweight multi-layer perceptron (MLP) architecture to outperform advanced CNNs and RNNs by achieving 99% accuracy at low SNRs (SNR -7) dB and >90% accuracy in moderate noise (SNR 4 dB). Additionally, this framework addresses the unexplored problem of simultaneous multi-source contamination (EMG+EOG+white noise), where it maintains 96% classification accuracy despite overlapping artifacts. With 30-second training times (97% faster than CNNs) and robust performance across SNR levels, this framework bridges the gap between clinical applicability and computational efficiency, which enables real-time use in wearable brain-computer interfaces. This work also challenges the ubiquitous dependence on model depth for EEG artifact detection by demonstrating that domain-informed feature fusion surpasses complex architecture in noisy scenarios.

**Keywords:** Electroencephalography, Noise classification, Ocular artifacts, Muscular artifacts, Deep neural network, Real-time processing

## 1. Introduction

Electroencephalography (EEG) is a non-invasive method for measuring the brain's electrical activity, that is widely used in clinical diagnosis, neuroscience research, and brain-computer interface (BCI) programs (Shih et al., 2012). However, the small amplitude of recorded EEG signals on the scalp makes them highly susceptible to external artifacts (signals that do not originate from the brain) (Minguillon et al., 2017). These artifacts commonly arise from electrical activities of the eyes, heart, muscles, and other interferences caused by cable movement and environmental electromagnetic noise (Jiang et al., 2019). In the absence of automated noise detection methods, operators identify noisy channels in EEG recordings manually. This process is time-consuming and depends on individual interpretation, which may result in inconsistent results depending on the expert evaluating the EEG (Daly et al., 2014; Debnath et al., 2020; di Fronso et al., 2019; Hartmann et al., 2014).

Noise removal methods range from widely known approaches like Independent Component Analysis (ICA) (Makeig et al., 1995) and multivariate empirical mode decomposition (MEMD) (Molla et al., 2012), Signal space reconstruction (Kothe & Jung, 2016), to more modern statistical techniques such as local outlier factor (Kumaravel et al., 2022) and deep learning-based approaches (Sun et al., 2020; Yang et al., 2018). These methods are applicable to multi-channel signals and therefore usually suffer from a high computational burden, which makes them inappropriate for real-time applications. Furthermore, while single-channel EEG is preferred for wearable BCIs and patient comfort (Acampora et al., 2021; Ko et al., 2017; Nguyen & Chung, 2018), existing single-channel solutions fail to address two prominent challenges: (1) simultaneous contamination by multiple artifacts (for instance EMG + EOG + white noise), and (2) the trade-off between accuracy and computational efficiency. For example, EEGDenoiseNet (Zhang, Zhao, et al., 2021) and subsequent CNNs (Paissan et al., 2022; Zhang, Wei, et al., 2021) focus on isolated artifacts like EMG or EOG, reaching low accuracy (less than 80% at SNRs higher than 2 dB). Additionally, the interpretable CNN shows reduced performance for mild noise at higher SNRs (above 2 dB). It is worth noting that previous approaches assumed that the data used for testing the designed network was a part of those that were used for training, but with different values of artifacts. This strategy may erroneously enhance the accuracy.

* Corresponding author, e-mail: smsakhaei@nit.ac.ir

The approach presented in this study overcomes previous limitations in noise detection methods for single-channel EEG signals, such as the high computational cost of multi-channel methods, user dissatisfaction, and the relatively low accuracy of earlier systems. This method employs some features derived in the time domain through low-pass filtering of the signal and in the frequency domain by power spectral density analysis, combined with a three-layer neural network for classification and detection. We have shown that our proposed method outperforms the existing deep learning-based systems both in accuracy and in other aspects.

## 2. Proposed Method

In this paper, we proposed a neural network with several fully connected layers for noise detection in single-channel EEG recordings. As shown in the flowchart in Fig. 1, our proposed model functions on spectral-temporal feature fusion extracted from a segment of EEG signals. These features are then processed using scaling and PCA techniques for computational efficiency before being fed into the network as inputs. Extracting appropriate features from EEG signals and performing steps such as component analysis play a pivotal role in enhancing the performance of machine learning systems. These features enable the network to extract all the crucial and characteristic information from the input data with ease. This section covers the preparation and preprocessing of data, introduces all the various features used and extracted in our proposed model, and addresses the role of each feature. Additionally, it offers a detailed analysis of the structure of the classifier along with the overall performance of the models.

### 2-1. Dataset

In this study, we utilized the semi-synthetic EEGDenoiseNet dataset (Zhang, Zhao, et al., 2021) to train and evaluate the proposed method performance. This dataset includes 4,514 clean EEG segments, 3,400 EOG segments (associated with eye blinks and movements), and 5,598 EMG segments (muscular artifacts). All segments were sampled at 512 Hz and a band-pass filter with a 1-80 Hz range was applied to all. These were then resampled to 256 Hz and divided into 2-second segments, each containing 512 samples. Fig. 2 illustrates the PSDs of randomly selected signals.

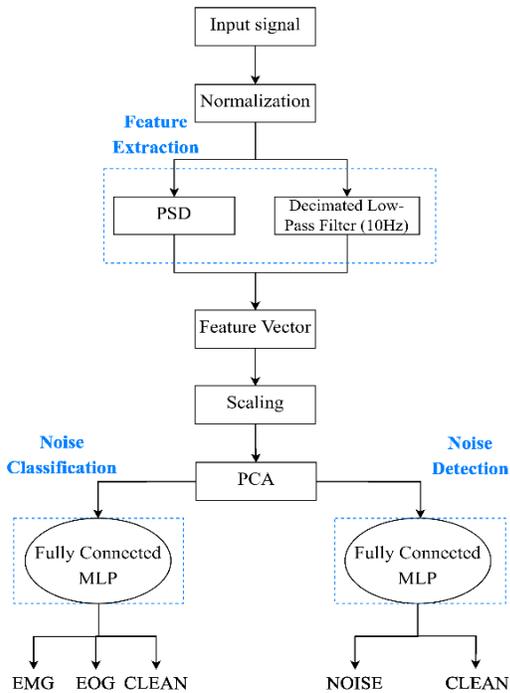

**Fig. 1** Flowchart of the proposed method

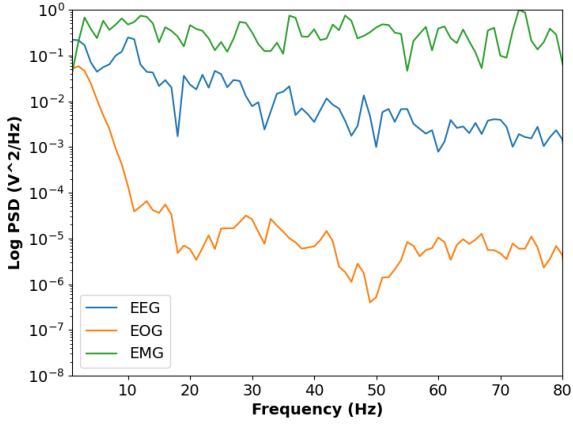

**Fig. 2** The PSD values of a segment randomly selected from EEGDenoiseNet

The noise segments, labeled as $\eta$ (EOG or EMG), and clean EEG segments, labeled as $x$, were combined linearly using Eq. (1) and Eq. (2) to produce a noisy signal $y$.

$$y_1(t) = x(t) + \lambda_{EOG}\eta_{EOG}(t) \qquad (1)$$

$$y_2(t) = x(t) + \lambda_{EMG}\eta_{EMG}(t) \qquad (2)$$

$$y(t) = \{x(t), y_1(t), y_2(t)\} \qquad (3)$$

where, $y_1$ and $y_2(t)$ are the signals contaminated by EOG and EMG noise, respectively, and $\lambda$ is the scaling factor parameter used for controlling the noise intensity, which is determined based on the Signal-to-Noise Ratio (SNR) as follows:

$$SNR[dB] = 20 \log \frac{RMS(x)}{RMS(\lambda\eta)} \qquad (4)$$

where $RMS$ denotes the root mean square, defined as:

$$RMS(g) = \sqrt{\frac{1}{N}\sum_{t=1}^{N}|g(t)|^2} \qquad (5)$$

## 2-2. Feature Extraction and Processing

### 2-2-1. Power Spectral Density (PSD)

PSD is a measure that provides the signal power distribution in the frequency domain. This feature assists the neural network in identifying critical frequency bands necessary for noise detection. Specifically, high-frequency bands such as beta (13-30 Hz) and gamma (30-120 Hz) waves in EEG, represent a critical overlap with EMG noise caused by muscle activity. In contrast, lower frequency bands like delta (1-4 Hz) and theta (4-8 Hz) are linked to eye movements and blinks, aiding in EOG noise detection.

Fig. 3 shows the PSD of a randomly selected segment contaminated with severe EMG and EOG noise (SNR: 7 dB), showing the differences between PSDs of clean, EMG-contaminated and EOG-contaminated signals. We used the Welch method (Welch, 1967) to estimate PSD with a 256 Hz sampling rate and Hanning window, a robust and widely used approach for EEG frequency analysis (Chai et al., 2015; Paissan et al., 2022; Zhang, Wei, et al., 2021). Welch method divides signals into overlapping sections, computes the Fast Fourier Transform for each and averages the results to produce a low-variance PSD estimate.

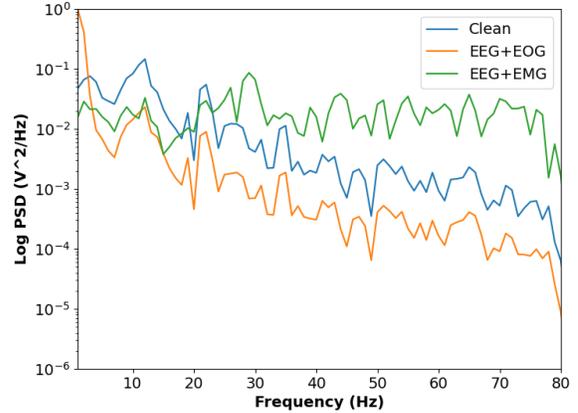

**Fig. 3** Log PSD values of noisy segments contaminated with EMG and EOG noises at SNR -7 dB

### 2-2-2. Low-Pass Filtering and Decimation

In this study, we employed a low-pass filter for extracting the low-frequency components of EEG signals, which is expected to be more relevant to EOG noise. We used a fourth-order Butterworth low-pass filter with a 10Hz cutoff frequency. The output was then decimated by a factor of 16 to effectively reduce the feature size from a 512-sample signal to 32 samples by capturing most of it.

## 2-3. Feature scaling

Standardization and scaling of features are the immediate next steps after feature extraction. Standardization involves scaling features in a way that the mean of each feature becomes zero, and its variance equals one. This process ensures that the classifier does not prioritize features with larger scales which allows all features to contribute equally during the training process. The standardization is performed as follows:

$$v\_scaled_i^j = \frac{v_i^j - \mu_j}{\sigma_j} \qquad (6)$$

where $v_i^j$ is the j-th feature value of $i$-th data sample, $\mu_j$ and $\sigma_j$ are the mean and standard deviation of $j$-th feature across the training dataset, respectively.

## 2-4 Principal Component Analysis

Principal Component Analysis (PCA) is a feature preprocessing method that is used before applying the features to the neural network, which can positively boost the performance. High collinearity among features can introduce challenges in training machine learning models and lead to reduced performance. In summary, using PCA enables the representation of data in a space with fewer features, simplifying temporal and spatial complexities (Qiu et al., 2024). We applied PCA on 112 extracted features (32 low-pass information and 80 PSD estimated samples). PCA extracted 73 components as shown in Fig. 4, cumulatively explaining 95% of the variance after applying PCA.

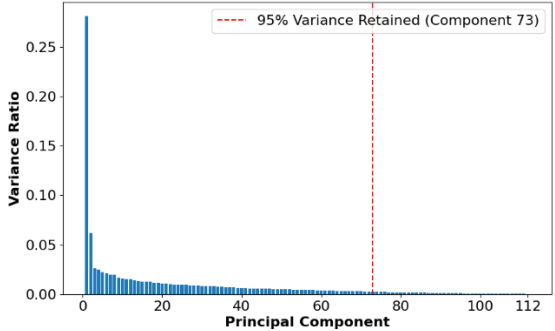

**Fig. 4** Variance ratio of principal components

## 2-5. Classifiers

The classifiers used here are simple multi-layer perceptron networks with three fully connected layers (excluding the input layer). For the binary classifier, the outputs represent the likelihood of the clean or noisy class and for the ternary classifier, the outputs represent the likelihood of each noise type. The weights for each layer are tuned while training to optimize classification accuracy. As shown in Fig. 5, the network architecture consists of a first hidden layer with 128 neurons and a second hidden layer with 64 neurons. Each hidden layer is followed by ReLU activation functions, which help introduce non-linearity and improve the learning process. The final output layer in the ternary classifier consists of three neurons, corresponding to the three target noise classes. In the binary classifier, the number of output neurons is reduced to two to classify the segment, whether to a clean or noise class. These neurons use the softmax activation function. The softmax function transforms the raw outputs of each neuron into probability values between 0 and 1, which ensures that the sum of these probabilities equals 1. In other words, the output of the last layer is a probability that the input signal belongs to each noise class. Additionally, a Dropout layer with a probability of 0.6 prevents overfitting. This layer randomly deactivates some of the 128 nodes before the fully connected layer during training. Due to its simplicity and efficiency, this structure is particularly effective for identifying different types of noise in EEG signals, especially when the noises are extracted using specific features introduced in this work.

The network is trained using the batch size of 32 and the cross-entropy loss function, which is helpful for multi-class classification problems and measures the variation between the model predictions and the actual labels. The Adam optimizer with an initial learning rate of 0.001 is employed due to its high convergence speed and effectiveness in training deep neural networks. The learning rate scheduler used is ReduceLROnPlateau from PyTorch (Paszke et al., 2024), which adjusts the learning rate based on the model performance on the validation dataset and reduces the learning rate when the validation loss plateaus, helping the model converge more efficiently. To further prevent overfitting, early stopping is implemented as well. As illustrated in Fig. 6, the early stopping effect can be seen in the loss function and accuracy based on the number of executed epochs. Out of 100 defined epochs, only a portion (around 40) is executed. This mechanism monitors the performance of the model on the validation dataset and halts the training process if no significant improvement in validation error is observed over 20 successive epochs. This approach helps avoid excessive training and unnecessary complexity, ensuring optimal accuracy.

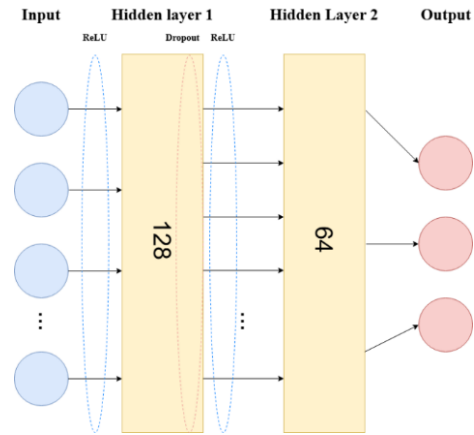

**Fig. 5** The general structure of the neural network (ternary classifier)

### 2.5.1– Training and Evaluation

Each clean EEG segment was combined once with muscle noise and once with ocular noise to create $2 \times 4512$ noisy segments. For 80% of the data (training and validation phases), $\lambda$ in Eqs. (1) and (2) were randomly selected to ensure diverse SNRs within a specified range. For the remaining 20% (test phase), 902 clean EEG segments, 902 noisy EOG segments and 902 noisy EMG segments were prepared. These were distributed into 13 separate directories, each containing 1,804 segments (902 noisy EOG and 902 noisy EMG), with SNRs ranging from -7 dB to 6 dB. For each directory, $\lambda$ was adjusted to achieve the desired SNR values for each segment. Before processing, all segments were temporally normalized to standardize their scale with zero mean and a unit standard deviation. This normalization eliminates physiological variations among individuals and recording conditions, easing comparisons and improving feature extraction by reducing undesirable fluctuations and balancing the signal. Normalization ensures extracted features represent more useful information from the signals.

## 3. Results

### 3.1- Binary Classifier (Noise detection)

As mentioned, a neural network architecture was developed to tackle two distinct but related tasks in single-channel EEG signals. First, turning to the noise detection task, Fig. 7 illustrates the binary network test results for noisy EEG contaminated with ocular and muscular artifacts for different values of SNR. Initially, the binary classification network performs almost perfectly for lower SNRs (below 2 dB) and stronger noises. However, a slight decline in the detection performance for EEG+EOG signals is observed compared to EEG+EMG as the noise becomes subtler. This is primarily because EOG contamination in clean signals is typically less pronounced and affects a narrower frequency range compared to EMG, and the diminishing amplitude of noises makes it more challenging for the network to detect them from the background signal. To analyze the models' performance, we present their confusion matrices in this section. As illustrated in Fig. 8, the confusion matrices for EMG and EOG detection at $SNR = 6\ dB$ show key intuitions of the model performance. For EMG detection, the classifier shows high precision, as it yielded only 2 false positives (not noisy, but detected as noisy) in opposition to 805 true positives (noisy and detected as noisy). However, the recall is slightly lower due to 97 false negatives, indicating that some noisy signals are misclassified as clean. EOG detection followed a similar pattern, with 4 false positives and 715 true positives. However, the recall is remarkably lower, with 187 false negatives.

An expected observation, however, is the reduction in accuracy for both tasks as SNR increases. This is due to noisy segments resembling clean segments at higher SNRs, making it harder for the network to distinguish between them. Unlike the ternary classifier, where the network must identify and distinguish between specific noise types (for example,

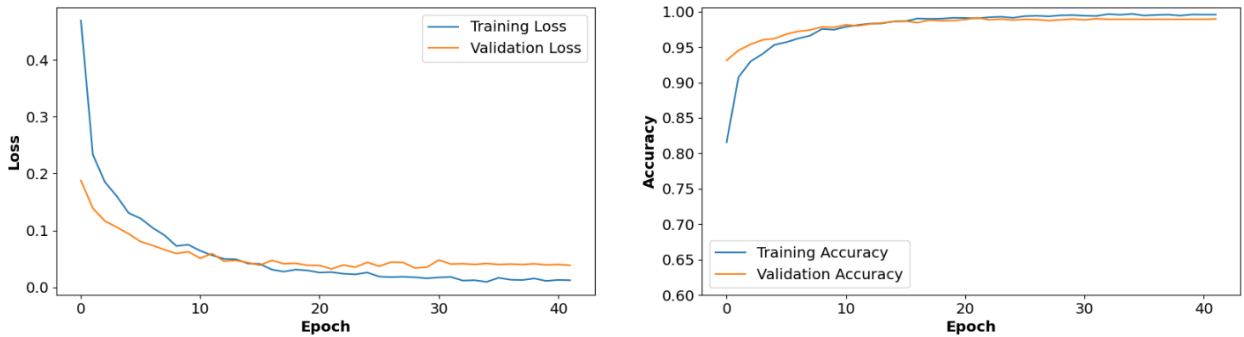

**Fig. 6** Loss and accuracy values during the training epoch of the ternary classifier

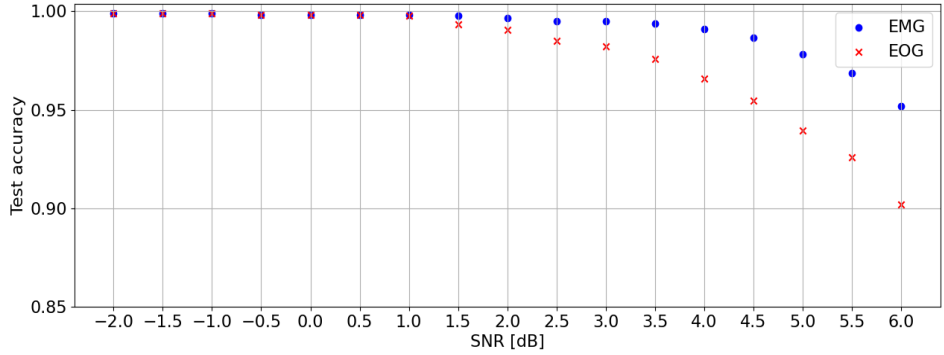

**Fig. 7** Detection accuracy of EEG contaminated by EMG and by EOG versus SNR

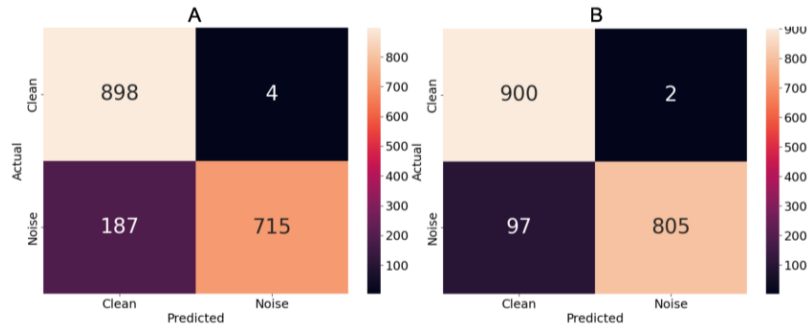

**Fig. 8** Confusion matrices for EOG detection (A), and for EMG detection (B) at SNR of 6 dB

**Table 1** Classification accuracy for different ternary classifiers at different values of SNR

| SNR (dB) | EEGDenoiseNet | 1D-CNN | EEGNet | SincNet | *Suggested M* |
|---|---|---|---|---|---|
| -7 | 0.65 | 0.98 | 0.98 | 0.99 | **0.99** |
| -5 | 0.65 | 0.98 | 0.97 | 0.99 | **0.99** |
| -2 | 0.65 | 0.97 | 0.96 | 0.99 | **0.99** |
| 0 | 0.65 | 0.96 | 0.92 | 0.99 | **0.99** |
| 2 | 0.64 | 0.88 | 0.85 | 0.86 | **0.98** |
| 4 | 0.64 | 0.75 | 0.75 | 0.75 | **0.96** |
| 6 | 0.64 | 0.70 | 0.70 | 0.75 | **0.90** |

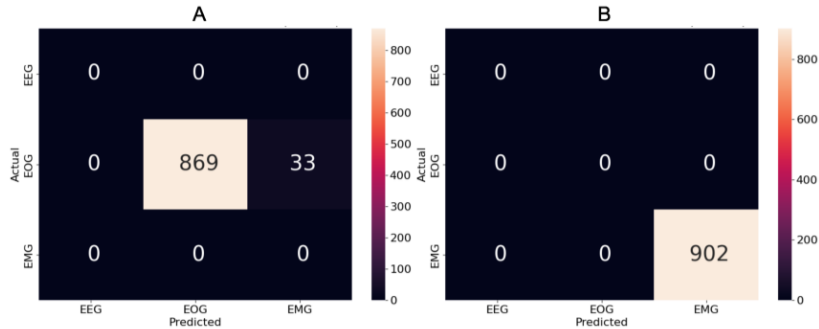

**Fig. 9** Confusion matrices for our ternary classifier in two situations: (A) Dominant ($SNR_{EOG} = -7dB$) with subdominant ($SNR_{EMG} = 2dB$) and ($SNR_{wn} = 3dB$) , (B) Dominant ($SNR_{EMG} = -7\ dB$) with subdominant ($SNR_{EOG} = 2\ dB$) and ($SNR_{wn} = 3dB$)

EMG or EOG), the binary classifier only determines whether a segment is noisy or clean. This task is intrinsically less complex and eliminates the need to categorize the noise; instead, it focuses on detecting the presence of any noise. Multiple simultaneous noise sources strengthen the classification of a segment as noisy. For example, when both EMG and EOG noises are present in a segment, their combined effect reduces the ambiguity associated with borderline cases, where the network might mislabel noise as clean or vice versa, especially at higher SNR where noise might otherwise appear subtle. Therefore, the binary classifier enhances its ability to distinguish noisy from clean EEG segments by utilizing overlapping noise features across frequency and time domains and this robustness is significantly important for real-world EEG signals, which makes the binary classifier an essential tool for preprocessing.

## 3.2- Ternary Classifier (Noise classification)

The ternary classifier extends the binary model by categorizing test segments into clean EEG, EMG-contaminated, or EOG-contaminated. This task is inherently more complex since overlapping spectral and temporal features of noise types, especially in real-world scenarios with mixed artifacts, entails more robust feature discrimination. The proposed method demonstrates higher noise detection and classification accuracy in a wide range of SNR levels, as shown in **Error! Reference source not found.**. At $SNR = -7\ dB$, the method achieves an accuracy of 0.99, comparable to state-of-the-art models like SincNet. Furthermore, as SNR increases, the accuracy remains consistently high, with values of 0.98 at 2 dB and 0.96 at 4 dB, while other methods, such as EEGNet and 1D-CNN experience a significant decline in performance. For instance, at SNR 4 dB, the suggested method accuracy is 96%, which is a 32% improvement over EEGDenoiseNet 64% accuracy and a 28% improvement over SincNet and other methods. Another significant finding of this research was the shorter training time of the proposed model due to its simpler structure compared to more complex models. **Error! Reference source not found.** represents the training times for different models during 50 epochs, which further demonstrates the efficiency of the proposed method, as it maintained a shorter training time (30 seconds) compared to more complex models like RNN-LSTM (346 seconds) and novel CNN architectures (1165 seconds) and highlighted the crucial importance of extracting appropriate features from EEG signals while maintaining a less complex architecture compared to others.

**Table 2** Time taken for training during 50 epochs

| Method | Training Time (s) |
|---|---|
| **Suggested Method** | **30** |
| **Interpretable CNN** | 43 |
| **fcNN** | 60 |
| **RNN LSTM** | 346 |
| **Simple CNN** | 496 |
| **Novel CNN** | 1165 |

This consistency is particularly advantageous for real-world applications where noise levels can vary, ensuring reliable performance in various scenarios. The gradual decrease in accuracy with increasing SNR, compared to the sharp drop observed in other models, suggests that the feature extraction and processing steps, including low-pass filtering, PSD analysis, and PCA, contribute to the robustness of the method.

## 3.3- Mixed scenario: a combination of different types of noises

In this study, we bridged a prominent gap in previous research by introducing a concurrent noise addition strategy to evaluate the robustness of the proposed network in real-world conditions instead of focusing on isolated or single-noise scenarios. Here, we assume that the EEG signal contains EMG, EOG and white noises, simultaneously and evaluate the performance of our proposed model. To this end, we consider one type of noise (EMG or EOG) to be dominant and the other one as subdominant. Furthermore, a constant power white noise was also introduced to all noisy segments. White noise, which is characterized by its uniform power across all frequency bands, was included to simulate environmental noise that is typically present in practical EEG, especially in wearable settings. For this simulation, we introduced the noises with constant SNR values of $SNR_{dominant} = -7\ dB$ for dominant noise, $SNR_{subdominant} = 2\ dB$ for subdominant noise, and finally, the white noise with the ratio of $SNR_{wn} = 3\ dB$. Fig. 9 shows the confusion matrices for these scenarios. The network achieves a high classification accuracy, correctly classifying 867 out of 902 instances of segments contaminated with dominant EOG, despite the presence of EMG and white noise. It seems that the overlapping characteristics of EMG and white noise at these SNR levels can occasionally confuse the model, particularly when the secondary noise approaches the SNR of the dominant noise. Nonetheless, the network successfully avoids misclassifying any EOG-dominated segment as clean EEG, which is crucial for real-world applications. Fig. 9 also shows the network ability to accurately classify segments dominated by EMG noise when $SNR_{EMG}$ = -7 dB, while $SNR_{EOG} = 2\ dB$ and $SNR_{WN} = 3\ dB$. Here, the model achieves perfect classification, correctly identifying 902 segments as EMG. This performance indicates that EMG characteristics and high-frequency components are distinct even in the presence of other noise sources which enables the model to differentiate EMG with high confidence. The addition of a constant power white noise ($SNR_{WN} = 3\ dB$) in both cases further indicates the robustness of the classifier. White noise, with approximately 10% of the dominant noise power, did not significantly affect the network ability to classify.

Results demonstrated that the network maintained robust performance despite the added complexity. The classification accuracy showed only a slight decrease, indicating that the feature extraction and processing methods effectively captured the distinguishing characteristics of the noise types, even when subdominant noise and white noise were present. These findings demonstrate the network potential for application in real-time noise detection tasks in complex, multi-noise environments and accentuate the effectiveness of the feature extraction methods in isolating and prioritizing dominant noise features even in complex noise environments and their effective operation in real applications. While the performance for EOG-dominated segments slightly lags behind that of EMG-dominated segments, the network overall robustness and high classification accuracy in real conditions make it a reliable tool for practical EEG noise detection and classification tasks.

## 4. Conclusion

In this study, we proposed a relatively lightweight and user-friendly approach for detecting and classifying noise in single-channel EEG signals, focusing on muscular (EMG) and ocular (EOG) artifacts. The user convenience is achieved by minimizing user input using single-channel data and simplifying the process, making it easier to apply in real-world scenarios. We developed two machine learning models: a binary classifier to identify whether an EEG segment is clean or contaminated by noise, and a ternary classifier to categorize the segments into Clean EEG, EOG or EMG. Features were extracted from both the time and frequency domains using a low-pass decimated Butterworth filter and Power Spectral Density (PSD) estimates, followed by dimensionality reduction via Principal Component Analysis (PCA). Additionally, the performance was evaluated under various noise conditions, including simultaneous contamination where both EOG and EMG are present, as well as white noise, and demonstrated robust results across a range of signal-to-noise ratios (SNRs).

The results of both our models revealed strong performance under various circumstances. The binary model classified noise-contaminated and clean EEG segments while maintaining an accuracy above 90% in SNRs ranging from -7 to 6 dB, with slightly lower confidence for EOG segments. The ternary classifier showed a similar pattern with the accuracy above 90% even in high SNR scenarios (above 2 dB) and maintained the high performance to stronger noise levels (down to -7 dB). This classifier also achieved 96% and 99% when subdominant and concurrent noises with around 10% of the dominant noise power were added to the segment. When compared to modern existing methods like RNN-LSTM and CNN, our model indicated both superior classification accuracy and faster training times, despite the simultaneous presence of subdominant noises and relatively less complex architecture, making it a more efficient deep learning solution for these tasks.

However, some challenges remain. While the method performed exceptionally well in controlled experiments, there were occasional misclassifications. For example, in the binary network and ternary network, it can be seen that the models had difficulties correctly identifying EOG. In the binary classification task, the classifier occasionally mislabels EOG as clean as SNR increases. In the ternary classifier, segments with dominant ocular artifacts were sometimes classified as EMG in cases where the intensities of EOG and EMG were similar or when white noise had a larger contribution to the signal contamination. This might potentially suggest that the disparity in the number of features used for each type of artifact failed to participate evenly in the model's final decision-making, as 80 PSD estimates are employed to detect EMG-contaminated segments while only 32 low-pass features are incorporated for EOG detection. Further investigation into this discrepancy is necessary to ensure that both EOG and EMG signals are equally well-represented in the feature extraction process.

In conclusion, our proposed approach significantly advances EEG noise detection and classification by combining accuracy, speed, and simplicity. While there are areas for further improvement, particularly in handling various noise types at varying SNR levels, this work represents an important step toward developing real-time, efficient, and effective noise removal methods for EEG signal processing.

## Information sharing statement

The code and dataset supporting the findings of this study are publicly available at:

- https://github.com/HosseinEn/EEG-Artifact-Detection
- https://github.com/ncclabsustech/EEGdenoiseNet

**Funding** The authors did not receive support from any organization for the submitted work.

**Code availability** The code supporting this study is available at https://github.com/HosseinEn/EEG-Artifact-Detection

**Data availability** This study used publicly available data from the EEGdenoiseNet repository (https://github.com/ncclabsustech/EEGdenoiseNet).

## Declarations

**Ethical approval** Not applicable.

**Competing Interests** The authors have no competing interests to declare that are relevant to the content of this article.

**Informed consent** Not applicable.

**Authors contribution** HE: Conceptualization, Methodology, Software, Formal analysis, Visualization, Writing–original draft. PJ: Conceptualization, Data curation, Methodology, Formal analysis, Writing–review & editing. SMS: Supervision, Project administration, Writing–review & editing.